\documentclass[letterpaper, 10 pt, conference]{ieeeconf}  % Comment this line out if you need a4paper

\pdfoutput=1                                              % Use this for arxiv
\IEEEoverridecommandlockouts                              % This command is only needed if 
\usepackage{cite}
\usepackage{amsmath,amssymb,amsfonts}
\usepackage{algorithmic}
\usepackage{graphicx}
\usepackage{caption}
\usepackage{subcaption}
\usepackage{textcomp}
\usepackage{xcolor}
\usepackage{siunitx}
\usepackage{float}
\usepackage{svg}
\usepackage{amsmath,amssymb}
\usepackage{verbatim}
\usepackage[normalem]{ulem}
\usepackage{todonotes}
\usepackage{soul}

\makeatletter
\let\NAT@parse\undefined
\makeatother
\usepackage{hyperref}

\graphicspath{ {images/} }

\def\BibTeX{{\rm B\kern-.05em{\sc i\kern-.025em b}\kern-.08em
    T\kern-.1667em\lower.7ex\hbox{E}\kern-.125emX}}

\title{\LARGE \bf Digital twins to alleviate the need for real field data in vision-based vehicle speed detection systems}

\author{A. Hern\'{a}ndez Mart\'{i}nez$^{1}$, I. Garc\'{i}a Daza$^{1}$, C. Fernández López$^{2}$ and D. Fern\'{a}ndez Llorca$^{1,3}$% <-this % stops a space
\thanks{$^{1}$ Computer Engineering Department, Polytechnic School, University of Alcal\'a, Madrid,  Spain. \{antonio.hernandezm,  ivan.garciad\}@uah.es \newline
$^{2}$ Institut für Mess- und Regelungstechnik (MRT), Karlsruher Institut für Technologie (KIT), Karlsruhe, Germany. carlos.fernandez@kit.edu
\newline
$^{3}$ European Commission, Joint Research Centre, Seville, Spain. david.fernandez-llorca@ec.europa.eu}%
}

\begin{document}

\maketitle
\thispagestyle{empty}
\pagestyle{empty}

\begin{abstract}
Accurate vision-based speed estimation is much more cost-effective than traditional methods based on radar or LiDAR. However, it is also challenging due to the limitations of perspective projection on a discrete sensor, as well as the high sensitivity to calibration, lighting and weather conditions. Interestingly, deep learning approaches (which dominate the field of computer vision) are very limited in this context due to the lack of available data. Indeed, obtaining video sequences of real road traffic with accurate speed values associated with each vehicle is very complex and costly, and the number of available datasets is very limited. Recently, some approaches are focusing on the use of synthetic data. However, it is still unclear how models trained on synthetic data can be effectively applied to real world conditions. In this work, we propose the use of digital-twins using CARLA simulator to generate a large dataset representative of a specific real-world camera. The synthetic dataset contains a large variability of vehicle types, colours, speeds, lighting and weather conditions. A 3D CNN model is trained on the digital twin and tested on the real sequences. Unlike previous approaches that generate multi-camera sequences, we found that the gap between the the real and the virtual conditions is a key factor in obtaining low speed estimation errors. Even with a preliminary approach, the mean absolute error obtained remains below 3km/h. 
%Accurate vision-based speed measurement is a method that has a great cost advantage over the traditional speed detection methods currently available. However, such systems also have their disadvantages due to the limitations of the cameras itself, which must be properly calibrated to give an accurate result. In most cases, as seen in \cite{Llorca2021}, they address these problems by focusing on systems based on deep learning, however, the main problem with this type of approach is that a large amount of data needs to be collected, among which both the traffic video sequences and the speed output data for each sequence must be found. As recently demonstrated in \cite{Hernandez2021} and \cite{Hernandez2022}, thanks to the use of simulators such as CARLA simulator, a sufficiently large and robust database can be generated, which can be used to train an architecture that, from different video sequences, is able to infer the speed of a vehicle on the road, with a simulation accuracy within the limits set by the different road safety institutions. In this work we propose the generation of a digital-twin of a real urban environment and, based on this digital-twin, train a network architecture capable of detecting the real speed of vehicles traveling on the road accurately, thus avoiding the step of setting up a whole system of image capture in a real urban environment.
\end{abstract}

\section{Introduction}
 \label{sec:introduction}
There is a clear causal link between speeding and safety risks, showing strong evidence that speed enforcement leads to a reduction of accidents \cite{ERSO2021}. Reducing speed limits in highways and rural roads improves the safety of all vehicle occupants, but in urban environments the positive impact also translates into a reduction in pedestrian fatalities \cite{Fridman2020}. Moreover, it has also proved to be beneficial for traffic noise, environment and health \cite{TE2021}. An increasing number of cities have implemented a 30km/h zone in the city centre. 

\begin{figure}[t]
    \centering
    \includegraphics[width=0.95\linewidth]{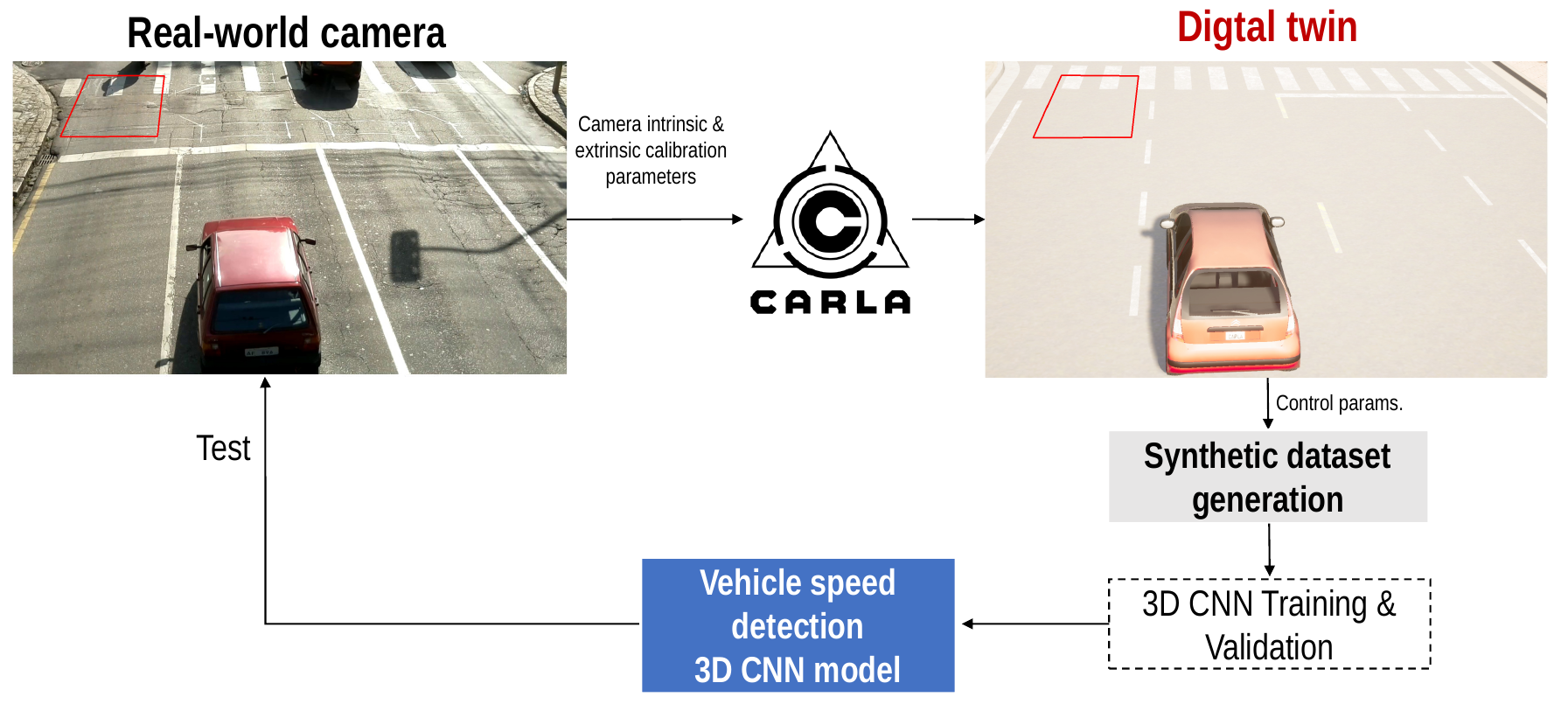}
    \caption{Overview of the presented approach. A digital twin is generated from the camera calibration parameters using CARLA simulator \cite{carla2017}. Then, a synthetic dataset is created and used to train and validate a 3D CNN model, which is directy used for speed detection in the real world.}
    \label{fig:overview}
    %\vspace{0mm}
\end{figure}

The enforcement of speeds limits plays a key role. Most studies confirm its positive influence, with estimated crash reductions of between $5\%$ and $69\%$ \cite{Pilkington2005}. Moreover, as the OECD stresses\cite{OECD2006}, drivers' expectations of being caught should extend beyond a few locations. The greater the number of enforcement points, the better the positive effects. However, as the requirements for accuracy and robustness in vehicle speed measurement for speed enforcement are very demanding \cite{Llorca2016}, the technology for speed detection usually involves high-accuracy, high-cost range sensor (radar, LiDAR or inductive loops). For municipalities, cost-efficient speed enforcement solutions are a necessary condition. 

The use of cameras and computer vision is beginning to be considered as a cost-effective alternative solution, with potentially enhanced functionality, for accurate speed estimation \cite{Llorca2021}. However, this a very challenging problem due to the discrete nature of video sensors, in which resolution decreases proportionally to the square of the distance \cite{Llorca2016} and the negative impact of adverse weather and lighting conditions. Unlike other problems where machine learning can be effectively applied, an added difficulty is the limited availability of data in real-world scenarios that allow the deployment of learning-based approaches. Data gathering in this area requires a complex and costly setup to capture images from cameras synchronised with some high precision speed sensor to obtain the ground truth values. The number of datasets with this type of information is still very limited, which implies that the use of data-driven strategies is far from being consolidated in this domain.

In our previous works \cite{Hernandez2021}, \cite{Hernandez2022}, we generated synthetic datasets from a driving simulator (CARLA~\cite{carla2017}) and used them to preliminary evaluate their feasibility for training and test deep learning models to perform speed regression from simulated sequences. In this work, we extend and validate this concept by generating a digital twin from a real environment (the UTFPR dataset \cite{Luvizon2017}). As showed in Fig. \ref{fig:overview}, the main idea is to create a synthetic replica of a real camera used for vehicle speed estimation by generating a sufficient number of synthetic sequences with known vehicle speeds and wide variability (speeds, lighting, weather, type of vehicles, etc.). Then we use this digital twin to train and validate a deep learning-based speed estimation model, which is finally applied the real environment. The results obtained are very promising, and allow us to envision a powerful methodology based on machine learning that does not require data captured in real environments. 

\section{Related work}
\label{sec:relatedwork}

As established in \cite{Llorca2021}, although we can find hundreds of works focusing on vision-based vehicle speed estimation, the topic is not sufficiently mature. A considerable number of works present sub-optimal camera pose and settings resulting in very high meter-to-pixel ratios that are unlikely to provide accurate measurements. Sensitivity to lighting and weather conditions, camera pose and settings are not sufficiently addressed. In addition, the number of learning-based approaches, while dominating in other domains, is still very limited for speed detection. This can be explained by the lack of consolidated datasets to train and compare the different methods. In this section, we focus on recent learning-based methods for vision-based vehicle speed detection from the infrastructure, including available datasets. We refer to the survey presented in \cite{Llorca2021} for a complete overview of the state-of-the-art.

\subsection{Learning-based approaches}
Average traffic speed estimation was posed as a video action recognition problem using 3D CNNs in \cite{Dong2019}, collating RGB and optical flow images. They emphasized that the main limitation was overfitting due to the lack of data. In \cite{Madhan2020} a Modular Neural Network (MNN) architecture was used to perform joint vehicle type classification and speed detection. In \cite{Revaud2021} camera calibration, scene geometry and traffic speed detection were jointly addressed by means of a transformer network trained with synthetic data, with the limited assumption that cars have similar and known 3D shapes with standardised dimensions. Although these three approaches are conceived for traffic speed detection, they can be adapted to perform single vehicle speed estimation.

Some works have proposed the use of recurrent architectures \cite{Parimi2021}, \cite{Hernandez2021}, \cite{Perunicic2023}. However, there is preliminary evidence showing a worse performance than with non-recurrent methods such as 1D CNN \cite{Cvijetic2023} or 3D CNNs \cite{Hernandez2021}. In \cite{RR2022}, after vehicle detection and tracking using YOLOv3 and Kalman filter, respectively, a linear regression model was used to estimate the vehicle speed. In \cite{Barros2021}, they used Faster R-CNN and DeepSORT methods to perform vehicle detection and tracking respectively. Next, they extracted dense optical flow using FlowNet2, and finally, used a modified VGG-16 deep network to perform speed regression. In \cite{Cvijetic2023}, a YOLOv5 detector was used to generate a 1D-feature vector based on the changing bounding box area of the vehicle. Then, they trained a 1D-CNN to perform speed estimation. 

\subsection{Datasets for vehicle speed detection}
As far as we know, only three datasets with real sequences and real speed values are publicly available so far. First, the \emph{BrnoCompSpeed} dataset \cite{Sochor2017}, which contains 21 sequences ($\sim$1 hour per sequence) with 1920 $\times$ 1080 pixels resolution images at 50 fps in highway scenarios. They obtained the actual speed values using a laser-based light barrier system. Second, the \emph{UTFPR} dataset \cite{Luvizon2017}, which includes 5 hours of sequences with 1920 $\times$ 1080 pixels resolution images at 30 fps in an urban scenario. They recorded the ground truth speeds using inductive loop detectors. Finally, a recent dataset combining video and audio data was presented in \cite{Djukanovic2022}, including 400 annotated sequences. However, they obtained the actual speed values using the on-board cruise control systems, which are neither sufficiently accurate nor homogeneous across models and makers. 

The limited number of available datasets with real field data is somewhat conditioning the use of synthetic datasets, which is becoming increasingly prevalent for this problem. For example, in \cite{Lee2019} a CNN model to estimate the average speed of traffic from top-view images is trained using synthesized images, which are generated using a cycle-consistent adversarial network (Cycle-GAN). Synthetic scenes with a resolution of 1024 $\times$ 768 pixels covering multiple lanes with vehicles randomly placed on the road are used in \cite{Revaud2021} to train and test the method used to jointly deal with camera calibration and speed detection. In \cite{Hernandez2021}, a publicly available synthetic dataset was generated using CARLA simulator, using one fixed camera at 80 FPS with Full HD format (1920 $\times$1080), with variability corresponding to multiple speeds, different vehicle types and colours, and lighting and weather conditions. This dataset was extended to include up to six different camera poses \cite{Hernandez2022} and a total of almost 3.7K sequences. We also found in \cite{Barros2021} a synthetic dataset using CARLA, including multiple cameras and generating more than 12K instances of vehicles speeds.

\section{Method}
 \label{sec:method}
This section describes the methodology used to generate a the digital twin based on the UTFPR dataset\cite{Luvizon2017},and the 3D CNN model used to perform speed estimation. %In this way, the construction of a digital-twin of a real environment is proposed, taking into account the camera calibration parameters, both  intrinsic and extrinsic (with respect to the road plane). %With this approach it is expected to be able to take the simulation results to an urban environment with real traffic.
Our approach aims to train a system from the digital twin only and then apply the trained model to the real urban environment without the need of adaptation.

\subsection{Digital twin generation}

\begin{figure*}[ht]
  \centering
  \includegraphics[width=0.90\linewidth]{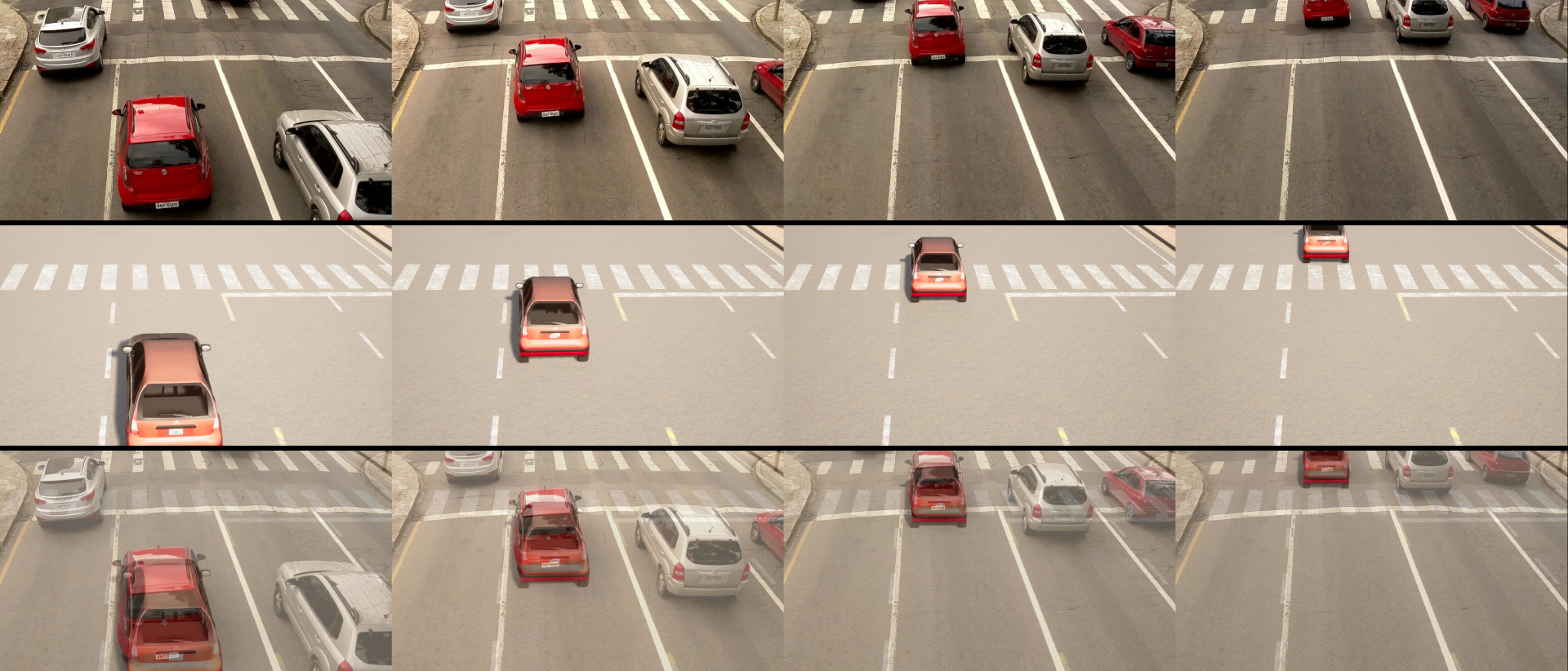}
  \caption{First row: Real sequence. Second row: Simulated sequence. Third row: Overlapped sequences.}
  \label{fig:OverlappedSeq}
\end{figure*}

%In order to generate the present dataset, it has been used the previous datasets generated in the previous works presented in \cite{Hernandez2021} \cite{Hernandez2022}. In this context, a digital twin of an existing real environment is generated, so it is necessary to be able to locate the camera in the best possible manner, knowing its extrinsic parameters.

\begin{figure}[!t]
    \centering    \includegraphics[width=0.5\linewidth]{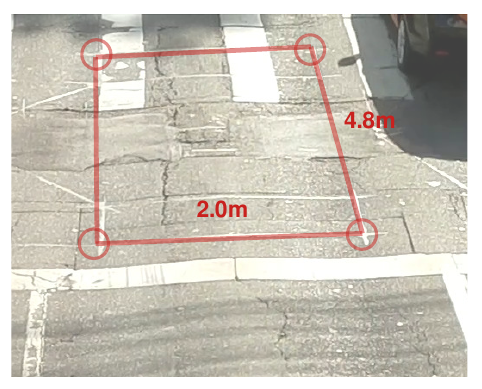}
    \caption{Dimensions of the inductive loops.}
    \label{fig:loop}
    %\vspace{0mm}
\end{figure}

The proposed digital twin simulates traffic on an urban road where environmental elements such as trees, zebra crossings, traffic signs, etc., as well as lighting and weather conditions are fully controlled. Images are rendered from the perspective of a virtual camera and used to build a dataset of traffic scenarios. The vehicle speed is a design variable, so it is known for any instant of time. The digital twin is based on CARLA driving simulator \cite{carla2017}, where the physics models of the actors are very accurate and depicts high-quality images in the rendering process due to using the Unreal-Engine graphic motor. The correct generation of the digital twin (see Fig. \ref{fig:OverlappedSeq}) has to consider two main factors. First, to replicate the intrinsic parameters of the virtual camera and the pose with respect to the road (i.e., the aspect ratio of a similar vehicle in real and simulation is maintained during the whole sequence). Second, to recreate with a certain degree of similarity the visual conditions of the simulation environment (it minimises possible system failures derived from components external to the studio's). Thus, the real system captures images of vehicles driving on the road with a fixed camera position: $[x, y, z, \alpha, \beta, \Omega]$, according to a system coordinate origin placed on the road plane. The camera's theoretical focal distance and optical centre (intrinsic parameters) used in the UTFPR dataset can be obtained from the camera model provided. However, the camera pose is unknown, except for the height, which has a value of $5.5$m. In addition, the authors also provide the size of the inductive loops, which are visible from the cameras (see Fig. \ref{fig:loop}) and where the world origin is placed, the upper left corner identified by the loop on the left, for pose camera evaluation. Therefore, optimisation techniques can be applied to estimate the camera pose. 

%\textcolor{red}{Image 1 shows how the Digital Twin is built, the obtained result and the spiral installed on the roadway to measure vehicle speed.  The spiral system measures the vehicle speed in that lane. This value is used as ground truth to evaluate precision metrics. The four corners of the loop are spaced $2m$ transverse to the road and $4.8m$ longitudinally to the street. }

As mentioned above, optimisation techniques are used to determine the camera pose since an analytical solution is complex to evaluate, the Stochastic Descent Gradient algorithm is used, and equation \ref{eq:loss} defines the cost function. 

\begin{equation}
    loss =  \sum_{1}^{4} \left\| x_i - \widetilde{x_i} \right\|_2 + \sum_{1}^{4} \left\| d_i - \widetilde{d_i} \right\|_2
    \label{eq:loss}
\end{equation}

In the equation, $x$ represents the coordinates of the corners of the [u,v] loop, while $d$ indicates the distance between two adjacent loop corners. 
%Figure \ref{fig:loss} illustrates the progression of time loss. 
The optimised camera pose is $[-3.613, 5.5, 19.567, 0.481, -0.059, -0.108]$ where the loss function value is around 78 pixels, considering the optics distortion the consequence of this.

%\begin{figure}[!t]
%    \centering
%    \includegraphics[width=.8\linewidth]{figures/loss.pdf}
%    \caption{Time loss evolution.}
%    \label{fig:loss}
%    %\vspace{0mm}
%\end{figure}

%%%%%%%

Since these parameters are not given in the dataset, a first approach has been focused on a multipose dataset, in which the camera pitch is varied for each sequence, in a range of 5 degrees [-33º,-37º]. 

%\textcolor{red}{Furthermore, another data set has been constructed using the camera intrinsics, which are known, and 4 points of the image, determined by the corners of the inductive loop present on the roadway, and whose coordinates, both real and pixel coordinates, are known.}

\begin{figure}[!t]
    \centering    \includegraphics[width=1.0\linewidth]{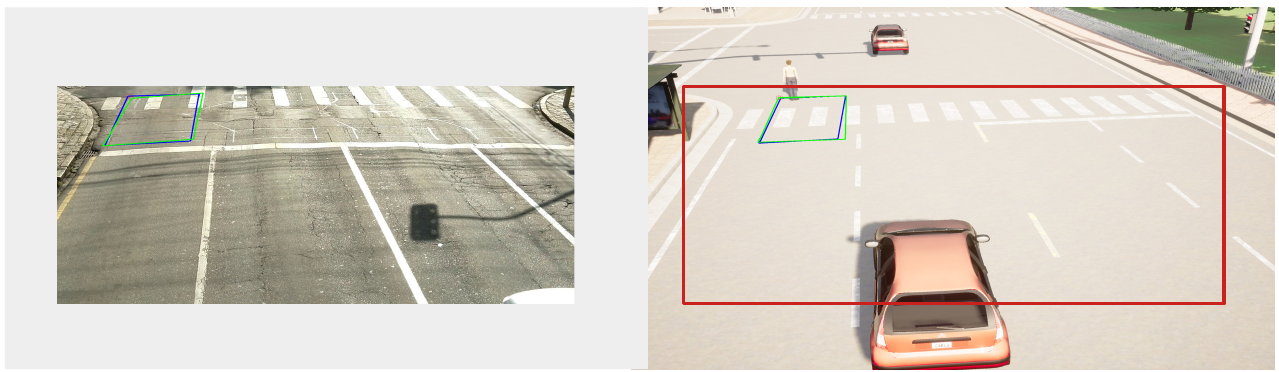}
    \caption{Digital Twin after the calibration process.}
    \label{fig:DigTwinCalib}
    %\vspace{0mm}
\end{figure}

We also remark that the original sensor image used to record the dataset had a resolution of 2592x1944, but configured to record at 30FPS with a resolution of 1920x1080 (after cropping). We generate the  images in CARLA with the same resolution and apply the same cropping. 

%\textcolor{red}{Using this information, it has been possible to perform an optimization process to find the extrinsic values of the camera in the real environment. Figure \ref{fig:OverlappedSeq} shows the final result of superimposing the real sequence with the simulated sequence. As can be seen, the position of the vehicle in simulation is quite accurate compared to its counterpart in the real environment.}

\begin{figure*}[ht]
  \centering
  \includegraphics[width=0.8\linewidth]{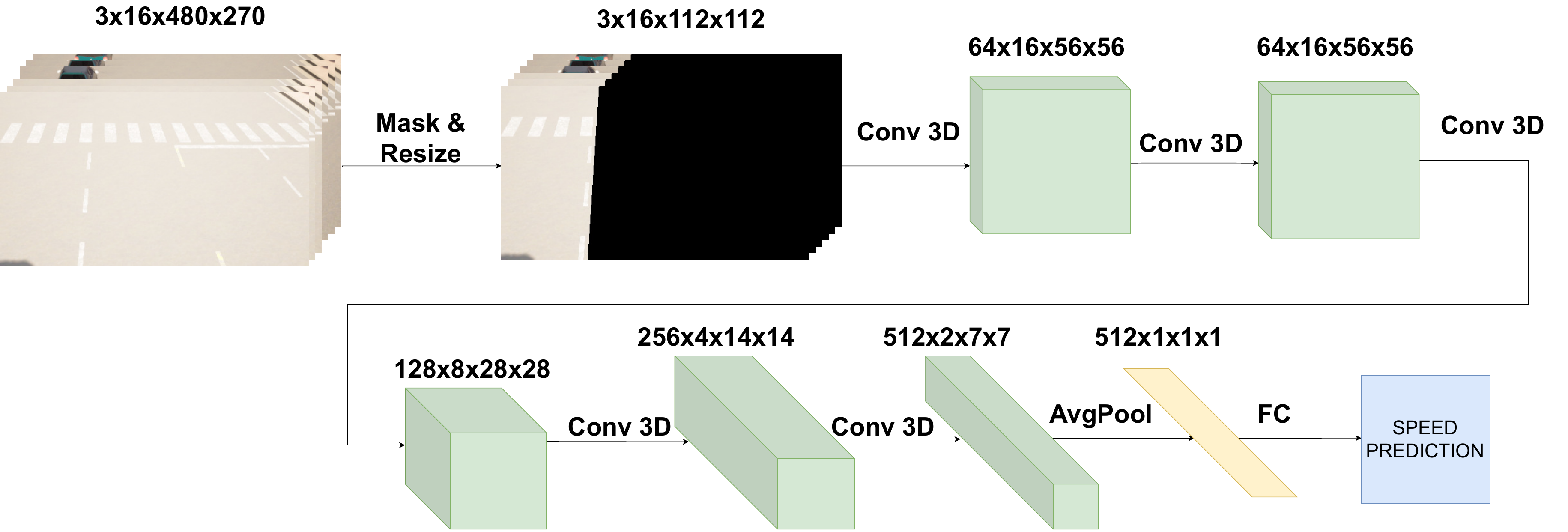}
  \caption{Overall view of the 3D CNN network architecture. The 4D input tensor contains a sequence of 16 RGB frames (3$\times$16) of 1.03 seconds duration, with image size of 112$\times$112 pixels. }
  \label{fig:resnet3d}
\end{figure*}

\subsection{Speed estimation}

\begin{figure}[!t]
    \centering
    \includegraphics[width=0.95\linewidth]{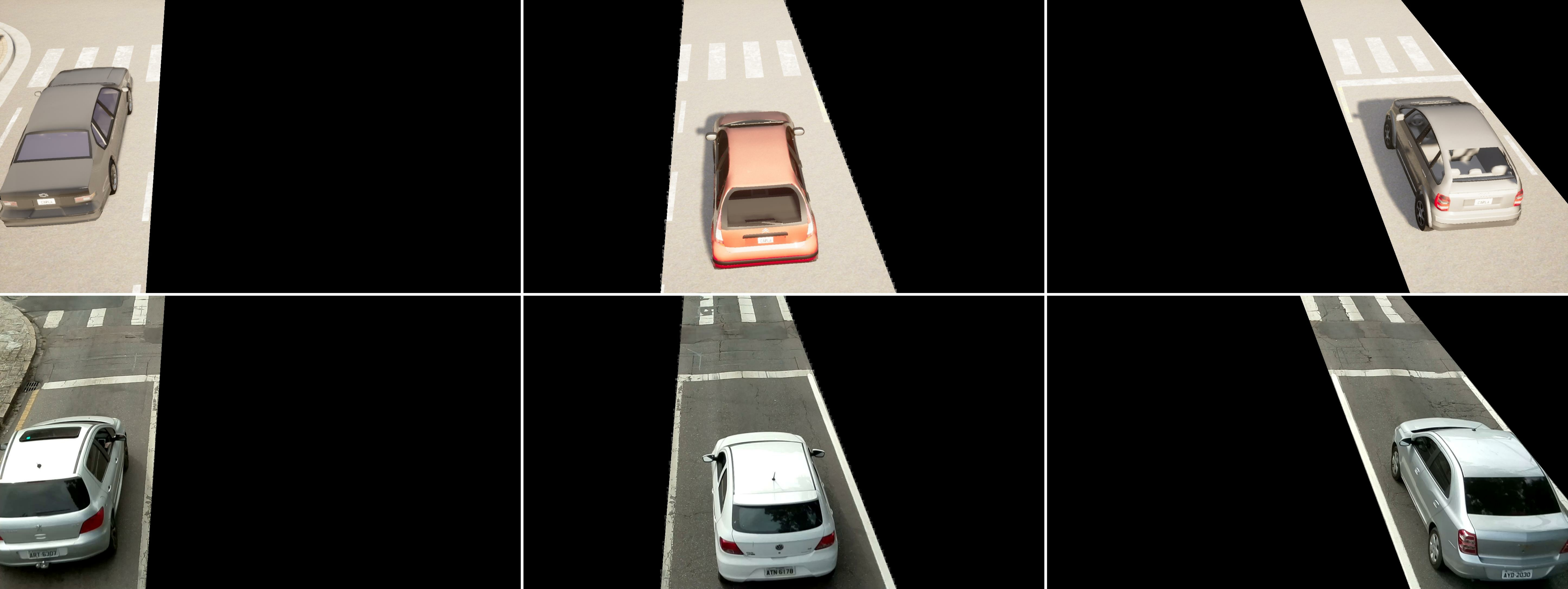}
    \caption{Overview of the image masking process per lane. Top row simulation, bottom row real environment.}
    \label{fig:masekdcollage}
    %\vspace{0mm}
\end{figure}

For this study, we have selected the ResNet 3D (3D CNN) architecture to perform speed regression. The overall architecture is depicted in Fig. \ref{fig:resnet3d}. This model has proven to be very effective in previous works \cite{Hernandez2021}, \cite{Hernandez2022} for lane-based vehicle speed detection using synthetic datasets. However, in the case of multi-lane/vehicle sequences (the UTFPR dataset camera covers up to 3 lanes) the input must be adapted so that the entire input sequence only contains data from one vehicle. A lane-based masking pre-processing is applied so that the model only processes one lane and one vehicle at a time (see Fig. \ref{fig:masekdcollage}). 

%For this study, the ResNet 3D (3D CNN) architecture (which has proven to be very effective in previous works) has been chosen again, since it has shown a better performance than the model using the feature extractor. Specifically, as could be seen in \cite{Hernandez2022}, it was tested with 3 sequences in real environment, which were recorded with a camera at different fps from the simulation recordings, and still achieved quite satifactory results. 

%%% TBC

%It is for this reason that the parameters and hyperparameters with which it was previously trained have been maintained. On this occasion, it has been decided to train a model with all the digital-twin sequences, in order to subsequently perform a test on the UTFPR dataset of traffic in a real environment. 

In order to evaluate the suitability of the proposed architecture to perform speed detection, we firstly performed a training, validation and test (60/20/20) using the real sequences and the ground truth speed values provided in the UTFPR dataset~\cite{Luvizon2017} (\textit{Real Environment}). 
%As we will see, the results were very accurate, which validate the proposed architecture for this particular environment. 
In addition, based on previous evidence \cite{Barros2021}, \cite{Hernandez2022}, we also studied the effect of using a multi-view synthetic dataset, generated with multiple virtual camera locations as in \cite{Hernandez2022} (\textit{Multi Camera}). Preliminary evidence suggested that a view-invariant speed detection system might be possible. However, as we will see, this hypothesis was not effective when applied to the real environment, as it generated a very large error. 

%First of all, the model has been trained with the digital-twin in which the camera position is variable, and in which the extrinsic and intrinsic parameters of the camera are not precisely calculated. This is done to check whether or not it is necessary to have the camera correctly calibrated, in order to take those parameters to simulation and try to replicate as accurately as possible the real environment.

%After this, a second training has been performed with the dataset in which the camera is positioned correctly, having calculated its extrinsics by optimization, and knowing its intrinsics by knowing the type of sensor used.

%Finally, these simulation-trained models have been used to perform a test on the total of the real UTFPR dataset. In this case, having used a completely different dataset for testing, it is guaranteed that the images of the real environment have never been previously seen by the neural network.

Furthermore, to explore the sensitivity to the initial position of the vehicle within the sequence used to train the speed detection model, we also created a synthetic dataset with the calibrated camera pose, but with vehicles starting earlier, so that the first frames contain the vehicle entering  the scene (\textit{Fixed Camera – Vehicle Earlier}). As we will see, the method was also largely affected by the initial vehicle position. 
Finally, we trained the 3D CNN model with a synthetic dataset generated with the calibrated camera view, and the same starting position for all the vehicles. The first frame of the sequence approximately corresponds to the first complete view of the vehicle, including the license plate (the method used in \cite{Luvizon2017} was based on the detection and tracking of the license plate). This model is what we call \textit{Digital Twin}, as it minimises the conditions between the virtual and the real environment. We also trained this model with a smaller version of the dataset (\textit{Digital Twin Small}). 

\section{Experimental Evaluation}
\label{sec:results}

\subsection{Training parameters} 
%The parameters and hyperparameters of the network have not been modified with respect to the previous work, since, as has been observed, good results were achieved with those used previously.
We used a learning rate of 3x10-4, a batch size of 5, the Adam optimizer, and the MSE as a loss function. The network is trained using early stopping to avoid overfitting. A normalization of the output speeds to the range [-1 ,1] has also been performed.

\subsection{Dataset size} 
In all datasets, we used a 60/20/20 distribution (training/validation/test). For the \textit{Real Environment} analysis, we use all the sequences available in the UTFPR dataset (7958). The previous \textit{Multi Camera} model was trained with a synthetic dataset containing 3660 sequences \cite{Hernandez2022}. For the \textit{Fixed Camera - Vehicle Earlier} and the \textit{Digital Twin} the datasets have a total of 6022 sequences. Finally, the \textit{Digital Twin (Small)} was created using 1800 samples. The range of speeds covered 10-80 km/h, and the type of vehicles used included different types of cars and motorbikes. 

%The training with the real dataset has also been performed to see the results obtained if the network was trained with a 100\% real model. In this case, the distribution for the real training is 1099 training samples (60\%), 368 validation samples (20\%) and 366 test samples (20\%).

%For the real and simulated datasets, the lanes have been masked since the tests previously performed with this architecture are prepared for a single lane, centered in the image, with a single vehicle per sequence. As this dataset is multi-lane (3 lanes of urban traffic) it is necessary to feed the network with sequences of a single lane and a single car per lane. Figure \ref{fig:masekdcollage} shows the masked sequences, both in real and in simulation.

\begin{figure*}[ht]
  \centering
  \includegraphics[width=0.95\linewidth]{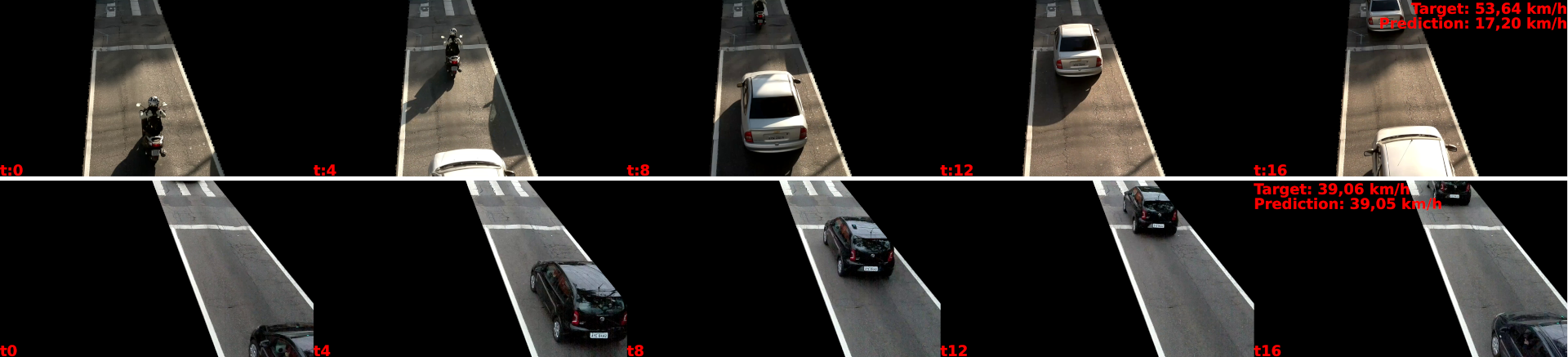}
  \caption{Top row: Sequence with highest error. Bottom row: Sequence with lowest error}
  \label{fig:MaskedResults}
\end{figure*}

\subsection{Results}

%The results obtained have been quite variable between the models trained with the multipitch dataset, the dataset with the camera in fixed position, and the real dataset. Firstly, with the training and test in real time, it has been observed that this type of methodologies, using a 3D network, behave correctly, yielding fairly accurate results, if there is a sufficient volume of data. The disadvantage of this is that it is difficult to find or generate a number of real environment sequences with accurate ground truth. This is the reason for trying to solve this kind of problem by generating synthetic datasets.

The obtained results are summarized in Table \ref{tab:ErrByDataset}. As can be observed, the proposed methodology proves to be very effective when training with real sequences, with a MAE of $0.53$ km/h. This value clearly outperforms previous works testing on the UTFPR dataset \cite{Luvizon2017}, \cite{Barros2021}, which is a contribution in itself. However, in our case, this only proves that the 3D CNN architecture is robust enough to deal with vehicle speed detection in this scenario, but our main objective is to avoid dependence on real data.

%After this, and as shown in Table \ref{tab:ErrByDataset}, the multipitch model has been discarded, as it shows that this type of system is very sensitive to camera position, and that a generic model with a variable set of camera positions does not return satisfactory results, compared to one in which the camera position is as close as possible to the real environment.

Despite preliminary evidence on the benefits of using synthetic sequences from multiple virtual cameras \cite{Barros2021}, \cite{Hernandez2022}, the results of the \textit{Multi Camera} model provided a very large error when applied to the real scenario (MAE $>26$ km/h). This suggests that the gap in the camera pose between the digital twin and the real-world scenario should be as low as possible. The sensitivity to the vehicle position within the input sequence is also a key factor, as the error obtained, even with the calibrated camera, when the position of the vehicles differ between the digital twin and the real environment is very large (MAE $>13$ km/h). 

Finally, when the vehicle position does not change between the synthetic and the real data, and the camera is properly calibrated, we obtained a reasonable error even with a small dataset (MAE of $3.03$ km/h). This error is reduced with a larger dataset (MAE of $2.66$ km/h), which also suggest that we may not have reached the optimal solution.

%Another source of error to take into account is that if the vehicle is placed inside the sequence from the beginning, as it is labelled in the UTFPR dataset, the error also drops considerably, proving that another source of error, besides the position of the camera, is the position of the vehicle along the sequence, which needs to be well adjusted.

\begin{table}[t]
    \centering
    \begin{tabular}{c|c|c}
        \hline
        \textbf{Dataset} & \multicolumn{2}{c}{\textbf{Mean Absolute Error [km/h]}} \\
    
        & Test In Sim.& Test In Real.\\ 
        
        \hline
        Real Environment &-&  0.532 \\
        Multi Camera & 0.25 & 26.35 \\
        Fixed Camera - Vehicle Earlier & 0.30 & 13.81 \\
        Digital Twin (Small) & 0.84 & 3.03 \\        
        Digital Twin (Large) & 0.74 & 2.66 \\
    \hline
    \end{tabular}
    \caption{Mean Absolute Error By Dataset.}
    \label{tab:ErrByDataset}
\end{table}

%beginning of the sequence, it can be seen that the error drops considerably and is within acceptable margins. This is also seen in Fig. \ref{fig:ErrBySpeed}. It is also normal for the error to increase due to the fact that in cases where the vehicle is travelling at high speeds, more vehicles enter the camera plane at the end of the sequence, as shown in Fig. \ref{fig:MaskedResults}, which can mislead the network.

\begin{figure}[t]
    \centering
    \includegraphics[width=.9\linewidth]{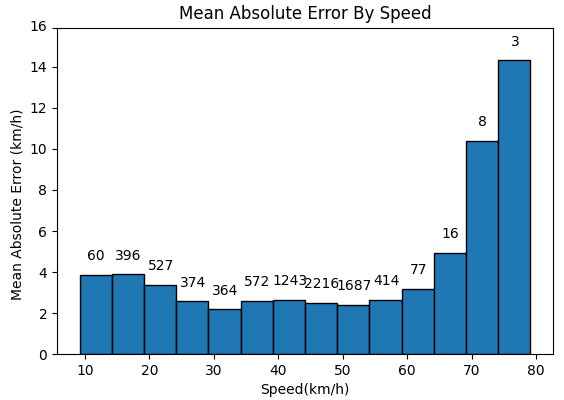}%{figures/19052023_MAE_BySpeed_CarInFirstImage.png}
    \caption{Error By Speed. The number on top of each bar is the number of samples for each range.}
    \label{fig:ErrBySpeed}
    %\vspace{0mm}
\end{figure}

\begin{figure}[t]
    \centering
    \includegraphics[width=.9\linewidth]{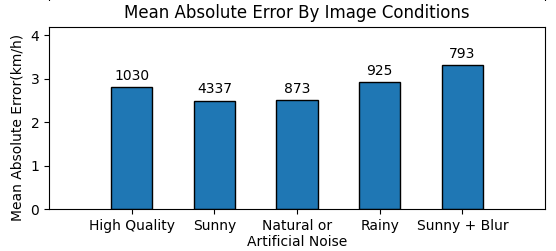}
    \caption{Error By Image Conditions. The number on top of each bar is the number of samples for each condition.}
    \label{fig:ErrByCond}
    %\vspace{0mm}
\end{figure}

The distribution of the MAE per speed is depicted in Fig. \ref{fig:ErrBySpeed}. The higher error for lower speeds can be explained by the fact that the scenario of the UTFPR dataset corresponds to an urban traffic light environment, and vehicles at low speed may be accelerating. The ground truth speed values are obtained from the inductive loops that are located far from the initial position. However, our approach includes samples far from the inductive loops, and the synthetic sequences have been generated with constant speed. 
Errors at higher speeds are consistent with most speed detection methods. In our case, this may be influenced by an insufficient number of images within the input 3D tensor. We have also analysed the error by lane (see Table \ref{tab:ErrByLane}). Although the difference is not significant, the Mean Absolute Error (MAE) is lower for the central lane than for the lateral ones, which in this case could be caused by a better calibration adjustment in the centre of the image.
Additionally, if we analyze the error by vehicle type (Table \ref{tab:ErrByVehicle}), it can be observed that the error for motorbikes is more than double that for cars. The smaller size of the motorbikes in the image, or their poorer realism in the simulated environment compared to reality, could be factors that negatively affect the results.
%On the other hand, other errors that have been observed are those due to accelerations, both in those cases of high speeds where the vehicle is braking, as well as those in which, from a red light, the vehicle must accelerate until it reaches the target speed. In this case, the ground truth of the dataset in real environment is calculated with inductive loops, which return the average speed between the 2 detections of the vehicle.
Finally, we can see the distribution of errors by image conditions in Fig. \ref{fig:ErrByCond}, including sunny and rainy conditions, as well as some images with artificial noise and some blurring. The results remain very stable regardless of the conditions. 

\begin{figure}[t]
    \centering
    \includegraphics[width=0.95\linewidth]{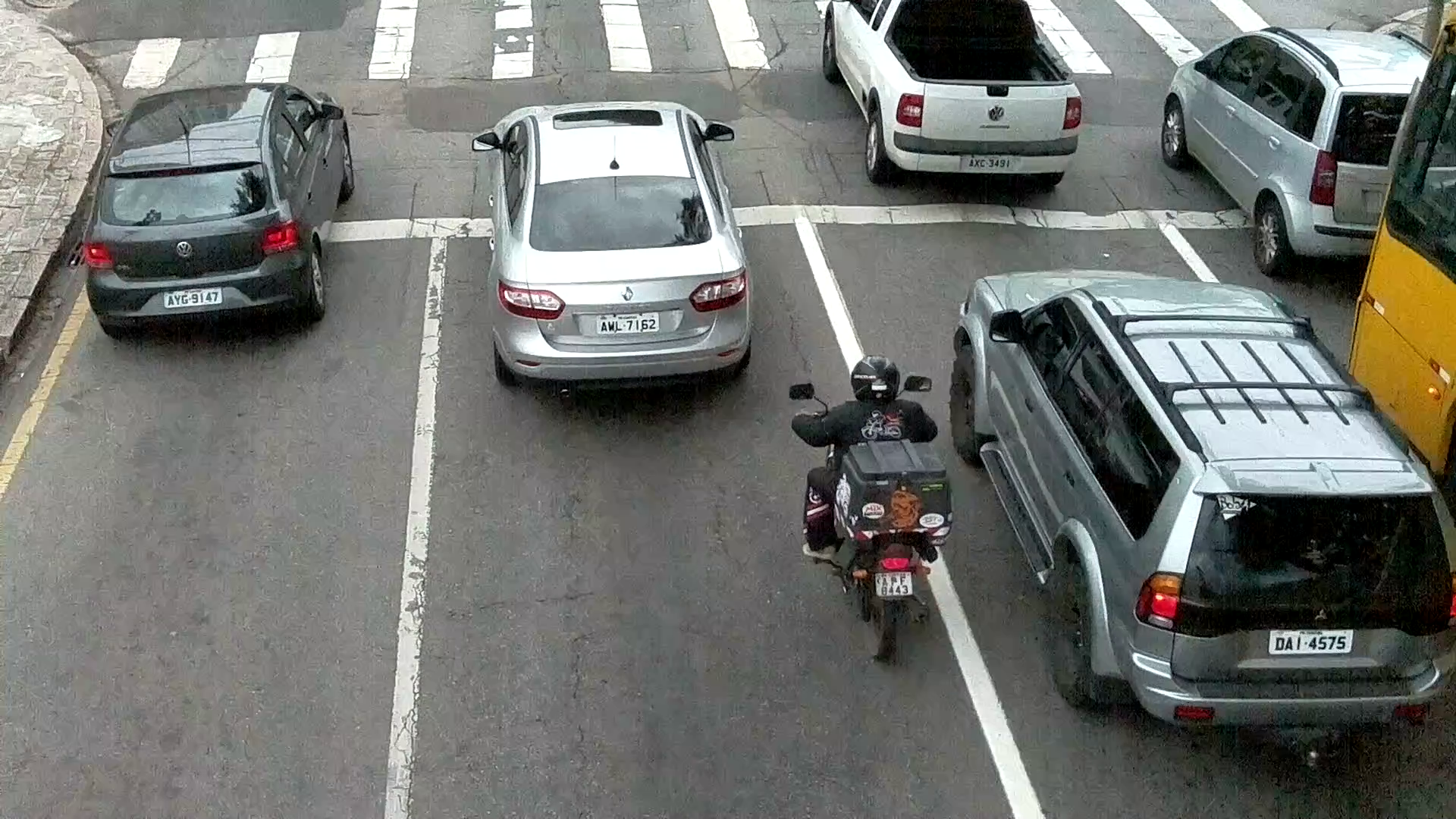}
    \caption{Two vehicles in the same sequence, central lane.}
    \label{fig:DoubleVehicle}
    %\vspace{0mm}
\end{figure}

If the samples with higher errors are analyzed, it can be observed that in these sequences more than one vehicle usually appears, as shown in Figs. \ref{fig:MaskedResults} (top row) and \ref{fig:DoubleVehicle}, in which it can be seen a motorcycle circulating very close to the car to be analyzed. Another fact to take into account is that the UTFPR dataset uses inductive loops as ground truth, so the model trained with the synthetic dataset inherently includes the error of these loops. 

%\begin{figure}[t]
    %\centering
    %\includegraphics[width=1\linewidth]%{figures/19052023_MAE_ByLane_CarInFirstImage.png}
    %\caption{Error By Lane. The number on top of each bar is the %number of samples for each range.}
    %\label{fig:ErrByLane}
    %\vspace{0mm}
%\end{figure}

\begin{table}[t]
    \centering
    \begin{tabular}{c|c|c}
        \hline
        \textbf{Lane} & \textbf{MAE [km/h]} & \# Samples \\
        \hline
        Left &2.50&  1616 \\
        Central & 2.36 & 3123 \\
        Right & 3.05 & 3219 \\
    \hline
    \end{tabular}
    \caption{Mean Absolute Error (MAE) and number of tested samples by lane.}
    \label{tab:ErrByLane}
\end{table}

\begin{table}[t]
    \centering
    \begin{tabular}{c|c|c}
        \hline
        \textbf{Vehicle} & \textbf{MAE [km/h]} & \# Samples \\
        \hline
        Car &2.60&  7786 \\
        Motorbike & 5.95 & 172 \\
    \hline
    \end{tabular}
    \caption{Mean Absolute Error (MAE)  and number of tested samples by vehicle type.}
    \label{tab:ErrByVehicle}
\end{table}

%Finally, the small digital twin returned a real error of 3.96 km/h, compared to an error of 2.89 km/h with the complete dataset. This proves the importance of sufficiently large and varied dataset to be able to get accurate results.  

%\vspace{-1}
\section{Conclusions}
 \label{sec:conclusion}

As has been demonstrated in this study, synthetic datasets for vehicle speed detection can be used to train a neural network, and then be able to transfer this model to the real world, with satisfactory results, without the need to provide the network with images in the real environment. This can bring many benefits, reducing costs, and avoiding the need of real data with associated ground truth speeds to deploy deep learning-based solutions. 

The sensitivity of the digital twin to the intrinsics and extrinsic parameters of the camera is considerable, so the gap between the virtual and real camera should be as low as possible. In addition, the position of the vehicles along the sequence should also be as similar as possible. The size of the dataset plays a key role, so a sufficient number of samples with sufficient variability is needed. 

Moreover, the proposed methodology allows a considerable number of future actions. We plan to continue expanding the dataset, with a larger number of samples and vehicles. In addition, it is also planned to carry out this type of study with other urban datasets with real traffic. In this case, to try to reduce the error, it would be necessary to filter the sequences avoiding cases of double detection by vehicles that are too close, and also take into account other error factors such as lane changes, acceleration, or  sequences including more than one vehicle. 
%in which before the end, there is another vehicle entering the image and can alter the results of the network. 
A possible solution to these problems could be to implement a vehicle detection system, based on deep learning, capable of tracking only the vehicle under study. 

\section{Acknowledgements}
This work was partially funded by Research Grants SBPLY/19/180501/00009 (Community of Castilla la Mancha),  PID2020-114924RB-I00 and PDC2021-121324-I00 (Spanish Ministry of Science and Innovation). D. Fernández Llorca acknowledges funding from the HUMAINT project by the Directorate-General Joint Research Centre of the European Commission. \underline{Disclaimer}: The views expressed in this article are purely those of the authors and may not, under any circumstances, be regarded as an official position of the European Commission.

\bibliographystyle{IEEEtran}
\bibliography{references}
\end{document}